\documentclass[conference]{IEEEtran}
\IEEEoverridecommandlockouts

\usepackage{cite}
\usepackage{amsmath,amssymb,amsfonts}
\usepackage{algorithmic}
\usepackage{graphicx}
\usepackage{textcomp}
\usepackage{xcolor}
\usepackage{multirow}
\usepackage{hyperref}
\usepackage[top=0.75in, bottom=1in, left=0.625in, right=0.625in]{geometry}

\def\BibTeX{{\rm B\kern-.05em{\sc i\kern-.025em b}\kern-.08em
    T\kern-.1667em\lower.7ex\hbox{E}\kern-.125emX}}
\begin{document}

\title{Hybrid Deep Learning for Hyperspectral Single Image Super-Resolution\\
}

\author{
    \IEEEauthorblockN{
        Usman Muhammad\textsuperscript{1} and Jorma Laaksonen\textsuperscript{1}
    }
    \IEEEauthorblockA{\textsuperscript{1} Department of Computer Science, Aalto University, Finland}
}


\maketitle

\begin{abstract}
Hyperspectral single image super-resolution (SISR) remains a challenging task due to the difficulty of restoring fine spatial details and preserving spectral fidelity across a wide range of wavelengths, which inherently limits the performance of conventional deep learning models. To effectively address this challenge, we introduce a novel module called Spectral-Spatial Unmixing Fusion (SSUF), which can be seamlessly integrated into existing 2D convolutional architectures to enhance both spatial resolution and spectral integrity. Specifically, the SSUF combines spectral unmixing and spectral–spatial feature extraction to subsequently guide a ResNet‑based convolutional neural network. Additionally, we employ a custom Spatial-Spectral Gradient Loss function, which integrates Mean Squared Error (MSE) with spatial and spectral gradient components, encouraging the model to accurately reconstruct features across both spatial and spectral dimensions. Experiments on three public remote sensing hyperspectral datasets demonstrate that our proposed hybrid deep learning (HDL) achieves competitive performance while reducing model complexity. The source codes are publicly available at: \href{https://github.com/Usman1021/hsi-super-resolution}{https://github.com/Usman1021/hsi-super-resolution}.
\end{abstract}

\begin{figure*}[!t]
\centering
\includegraphics[width=0.78\textwidth]{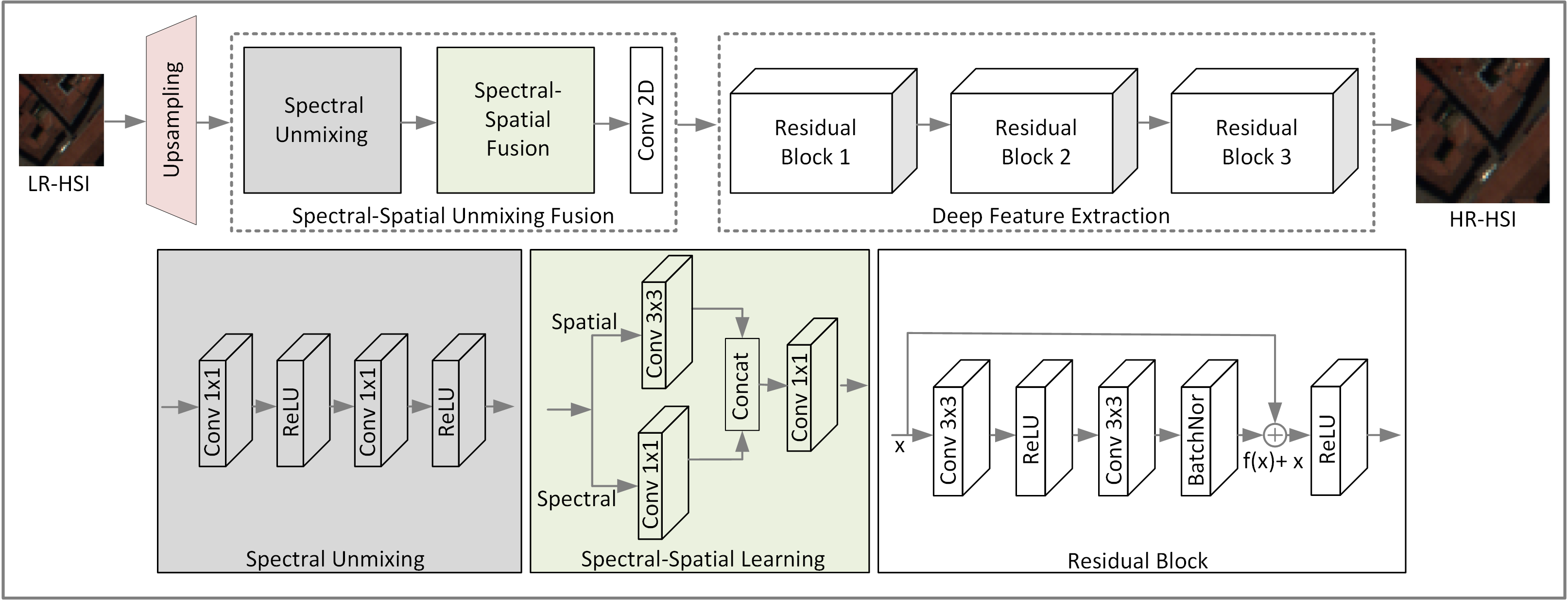}
\caption{Overview of the proposed hybrid deep learning (HDL) model. The gray block represents the Spectral Unmixing module, while the light green block denotes the Spectral–Spatial Learning module, which jointly extracts and fuses spatial and spectral features. Together, they form the Spectral–Spatial Unmixing Fusion (SSUF) block. The output is passed through a convolutional layer and then processed by the Residual Learning blocks (shown in white).}
\label{fig1}
\end{figure*}

\section{Introduction}
Hyperspectral images (HSIs) comprise hundreds of spectral bands, with a typical spectral resolution of approximately 10 nm ~\cite{hu2021hyperspectral, muhammad2018pre, muhammad2018feature, muhammad2019bag, muhammad2022patch}. This fine spectral granularity enables the detailed characterization of material properties, facilitating applications such as vegetation monitoring, mineral exploration, and environmental change detection ~\cite{obermeier2019grassland}. However, due to the limited energy available during image capture by hyperspectral sensors, balancing spectral resolution, spatial resolution, and signal-to-noise ratio (SNR) is essential to effectively capture the full details of the objects \cite{bu2024sbhsr}. Thus, to improve spatial super-resolution, several approaches can be employed: (1) fusion-based methods that combine HSIs with auxiliary images; (2) sub-pixel-based analysis; and (3) single-image super-resolution (SISR) methods \cite{mei2017hyperspectral}. The first two methods have been commonly used in hyperspectral applications. Specifically, fusion-based methods typically integrate low-resolution (LR) HSI with high-resolution (HR) panchromatic, RGB, or MSI images to produce high-spatial-resolution target images \cite{mei2017hyperspectral}. However, fusion-based methods rely on the availability of a high-resolution, co-registered auxiliary image to enhance the spatial resolution of HSIs \cite{li2023rgb}. This limitation restricts practicality in real-world applications.

In contrast to fusion-based methods, sub-pixel-based analysis focuses on spectral unmixing, which involves decomposing mixed pixels into their pure spectral components (called \textit{endmembers}) and their corresponding proportions (called \textit{abundances})~\cite{keshava2002spectral}. In particular, spectral unmixing has been explored in various forms, including linear spectral unmixing~\cite{lanaras2017hyperspectral}, nonlinear unmixing~\cite{zhao2021hyperspectral}, and deep learning–based unmixing~\cite{zhao2024ae}. Since natural hyperspectral images typically exhibit strong correlations in both spatial and spectral domains, effectively leveraging this prior information is crucial for enhancing unmixing performance~\cite{zhao2021plug}. 

Recently, convolutional neural networks (CNNs) have been widely used for SISR~\cite{dong2015image}, where spectral–spatial feature extraction has become a key component in many deep learning architectures, including dual graph autoencoders~\cite{zhang2022spectral} and Transformers \cite{lu2023spectral}, enabling the modeling of complex spectral correlations. However, SISR methods, which primarily focus on local regions, often struggle to capture non-local dependencies. These are essential for modeling global context and long-range spectral correlations in hyperspectral images~\cite{zhao2021plug, muhammad2025dacn}. Another challenge is that previous SISR models often process spatial and spectral information separately or fuse them only at later stages in the pipeline, which risks losing important spatial–spectral correlations~\cite{hu2021hyperspectral, li2023multiscale}.

To address the aforementioned challenges, we introduce a hybrid module called Spectral–Spatial Unmixing Fusion (SSUF), which combines spectral unmixing and spatial–spectral feature extraction to effectively exploit both local and non-local spectral–spatial information. Specifically, SSUF integrates spectral unmixing with spectral–spatial feature extraction at an early stage, encouraging joint learning of spatial edges and spectral relationships. In particular, the spectral unmixing component is designed to focus purely on spectral feature transformation for each pixel independently, allowing the network to learn nonlinear relationships across spectral bands. By doing so, the proposed scheme does not rely on a fixed number of endmembers~\cite{lanaras2015hyperspectral}. 

Meanwhile, the spectral–spatial feature extraction strategy is employed to jointly exploit both spatial and spectral information. This enables the network to capture local spatial context while preserving the spectral discriminability of the data. The output feature maps from the spectral unmixing and spectral–spatial learning modules are first passed through a convolutional layer and then fed into ResNet-like convolutional blocks for deep feature extraction and further refinement. Finally, we employ a custom spatial–spectral gradient loss function, which combines MSE loss with spatial and spectral gradient losses, guiding the model to accurately reconstruct gradients along both spatial and spectral dimensions.

In summary, our contributions are three-fold:

\begin{enumerate}
    \item We propose a novel SSUF module, designed for 2D deep learning architectures, to enhance both spatial resolution and spectral integrity.
    \item We employ a custom spatial–spectral gradient loss function that combines MSE with spatial and spectral gradient, enabling accurate reconstruction of gradients across both spatial and spectral dimensions.
    \item Experiments on three hyperspectral datasets under $2\times$, $4\times$, and $8\times$ downsampling scenarios demonstrate highly competitive performance across all datasets.
\end{enumerate}

\section{Methodology}
Our proposed hybrid deep learning (HDL) model, shown in Fig. 1, consists of two main components: (1) the Spectral–Spatial Unmixing Fusion (SSUF) block and (2) the Residual learning blocks. We begin by describing the first component, which includes spectral unmixing and spectral–spatial feature learning, followed by description of the residual blocks. Finally, we present the proposed custom loss function, which combines MSE with a spatial–spectral gradient loss to guide the overall training process.

\subsection{Spectral Unmixing}

The spectral unmixing composed of two successive $1 \times 1$ convolutional layers with ReLU activation, and is designed to focus purely on spectral feature transformation for each pixel independently. In particular, the double-layer structure enhances expressive power while maintaining computational efficiency. Mathematically, given an input tensor $X \in \mathbb{R}^{H \times W \times C}$, where $H$, $W$, and $C$ denote height, width, and number of spectral bands, respectively, the output of the Spectral Unmixing is given by:

\begin{equation}
U = \phi\left( W_2 * \left( \phi(W_1 * X) \right) \right)
\end{equation}

\noindent
where $W_1 \in \mathbb{R}^{1 \times 1 \times C \times F}$ and $W_2 \in \mathbb{R}^{1 \times 1 \times F \times F}$ are the learnable weights of the two convolutional layers, $U$ is the number of output feature channels, and $\phi(\cdot)$ denotes the activation function (e.g., ReLU).

\subsection{Spectral-Spatial Learning}

The spectral-spatial fusion is designed to jointly capture spatial and spectral features. It processes the input in two parallel branches. Specifically, a spatial branch using a $3 \times 3$ convolution to capture local spatial context while a spectral branch using a $1 \times 1$ convolution to transform spectral signatures. The outputs from both branches are concatenated and passed through an additional $1 \times 1$ convolution for fusion. Formally, the output $Y$ is computed as:

\begin{equation}
Y = \phi\left( {Q}_{f} * \left[ \phi({Q}_{s} * {X}) \,\|\, \phi({Q}_{\lambda} * {X}) \right] \right)
\end{equation}

\noindent
where ${X} \in \mathbb{R}^{H \times W \times C}$ is the input feature tensor, ${Q}_{s}$: learnable weights of the $3 \times 3$ spatial convolution, ${Q}_{\lambda}$: learnable weights of the $1 \times 1$ spectral convolution, ${Q}_{f}$: learnable weights of the $1 \times 1$ fusion convolution.

\subsection{Residual Learning}
Recently, residual networks demonstrated excellent performance in addressing the challenges of hyperspectral image super-resolution processing, particularly in single image super-resolution \cite{lim2017enhanced}. Thus, the residual blocks used in our model also follow the classical ResNet design. It consists of two convolutional layers with a shortcut (identity) connection that adds the input to the output of the second convolution. The block also includes a Batch Normalization layer and ReLU activations to improve learning dynamics and non-linearity.

Given an input tensor ${X} \in \mathbb{R}^{H \times W \times C}$, the residual block produces the output ${Z}$ using the following formulation:

\begin{equation}
{Z} = \sigma \left( \lambda  \left( {R}_2 * \sigma( {R}_1 * {X}) \right) + {X} \right)
\end{equation}

\noindent
where ${R}_1$ and ${R}_2$ are the learnable weights of the two $3 \times 3$ convolutional layers, $\sigma(\cdot)$ denotes the activation function (e.g., ReLU), $\lambda(\cdot)$ represents the Batch Normalization, ${X}$ is the input to the block and also serves as the identity shortcut connection.

\begin{table}
\caption{Ablation results of quantitative performance on the PaviaU dataset at scale 4 with model complexity}
\label{tab:ablation}
\centering
\resizebox{0.9\columnwidth}{!}{
\begin{tabular}{c|c|c}
\hline
\multicolumn{3}{c}{Ablation Study on PaviaU ($4\times$)} \\
\hline
Model Variant & MPSNR$\uparrow$ & SAM$\downarrow$ \\
\hline
Residual blocks + MSE loss & 29.84 & 5.31 \\
\hline
Residual blocks + Band grouping + MSE loss  & 30.51 & 4.61 \\
\hline
Spectral unmixing + Residual blocks + MSE loss  & 30.60 & 4.49 \\
\hline
Spectral-spatial learning  + Residual blocks + MSE loss  & 30.64 & 4.58 \\
\hline
Spectral–spatial unmixing fusion (SSUF) + MSE loss & 30.68 & 4.61 \\
\hline
SSUF with four Residual blocks + Custom loss & 30.51 & 4.70 \\
\hline
SSUF + Inception \cite{muhammad2025fusion} & 30.69 & 4.57 \\
\hline
SSUF + MobileNet \cite{muhammad2025towards} & 30.57 & 4.61 \\
\hline
SSUF +  Residual blocks +  Custom loss  (Ours) & \textbf{30.73} & \textbf{4.54} \\
\hline
\multicolumn{3}{c}{} \\ 
\hline
\multicolumn{3}{c}{Model Complexity} \\
\hline
Model & Scale & Parameters \\
\hline
ERCSR \cite{li2021exploring}         & 4 & 1.59M \\
\hline
MCNet \cite{lii2020mixed}             & 4 & 2.17M \\
\hline
PDENet \cite{hou2022deep}            & 4 & 2.30M \\
\hline
CSSFENet \cite{zhang2024hyperspectral} & 4 & 1.61M \\
\hline
FGIN \cite{muhammad2025fusion}    & 4   & 1.07M \\
\hline
DSDCN \cite{muhammad2025towards}    & 4 & 0.96M \\
\hline
HDL (Ours)                           & 4 & \textbf{0.33M} \\
\hline
\end{tabular}
}
\end{table}

\subsection{Spatial-Spectral Gradient Loss}
The spatial-spectral gradient loss is the total loss function of three components: reconstruction loss, spatial gradient loss, and spectral gradient loss. It is defined as:
\begin{equation}
L_{\text{total}} = L_{\text{MSE}} + \lambda_{\text{spatial}} L_{\text{spatial}} + \lambda_{\text{spectral}} L_{\text{spectral}},
\end{equation}
where \(L_{\text{MSE}}\) is the reconstruction loss, \(L_{\text{spatial}}\) is the spatial gradient loss, and \(L_{\text{spectral}}\) is the spectral gradient loss, with \(\lambda_{\text{spatial}}\) and \(\lambda_{\text{spectral}}\) as balancing weights. Specifically, the reconstruction loss (MSE) between \(z_{\text{true}}\) and \(z_{\text{pred}}\) is:
\begin{equation}
L_{\text{MSE}} = \frac{1}{N} \sum_{x,y,b}^{} \left( z_{x,y,b}^{\text{true}} - z_{x,y,b}^{\text{pred}} \right)^2,
\end{equation}
where \(N\) is the number of pixels across all spectral bands, and \( x,y,b\) represents horizontal, vertical, and spectral coordinates. Similarly, the spatial gradient loss is the sum of the horizontal (\(\nabla_x\)) and vertical (\(\nabla_y\)) squared gradient differences \cite{zhang2024hyperspectral}:
\begin{equation}
\small
L_{\text{spatial}} = \frac{1}{N} \sum_{x,y,b} \left[ \left( \nabla_x z_{\text{true}} - \nabla_x z_{\text{pred}} \right)^2 + \left( \nabla_y z_{\text{true}} - \nabla_y z_{\text{pred}} \right)^2 \right],
\end{equation}
where gradients are computed as:
\begin{equation}
\nabla_x z = z(x+1, y, b) - z(x, y, b),
\end{equation}
\begin{equation}
\nabla_y z = z(x, y+1, b) - z(x, y, b).
\end{equation}
Finally, the spectral gradient loss penalizes spectral deviations as:
\begin{equation}
L_{\text{spectral}} = \frac{1}{N} \sum_{h,w,b} \left( \nabla_b z_{\text{true}} - \nabla_b z_{\text{pred}} \right)^2,
\end{equation}
where the spectral gradient is:
\begin{equation}
\nabla_b z = z(x, y, b+1) - z(x, y, b).
\end{equation}

\section{Experimental setup}

\subsection{Datasets}
The experiments make use of three publicly available hyperspectral datasets to assess the performance of our model. These include the Chikusei dataset, the Pavia center (PaviaC) dataset, and the Pavia university (PaviaU) dataset.  The Chikusei dataset comprises 128 spectral bands, whereas the PaviaC and PaviaU datasets contain 102 and 103 spectral bands, respectively.

\subsection{Implementation and Evaluation Metrics}
To train and test the model on the PaviaC and PaviaU datasets, we used a patch size of $144 \times 144$, following the protocol of previous work \cite{zhang2024hyperspectral}. For the Chikusei dataset, we used a patch size of $128 \times 128$, as defined in \cite{wang2024enhancing}, with a center crop size of $512$. To convert the images to low resolution, the extracted patches are downsampled using area-based interpolation with scale factors of $2\times$, $4\times$, and $8\times$. During training, the Adam optimizer was used with a batch size of $4$. The weights were set to $2.0$ for the MSE component, $0.1$ for the spatial gradient, and $0.1$ for the spectral gradient. In addition, an early stop function was applied to prevent fixed epochs and avoid overfitting. Pre-upsampling was applied using bilinear interpolation. We adopted several widely used metrics to evaluate the quality of reconstructed images, including mean peak signal-to-noise ratio (MPSNR), Spectral Angle Mapper (SAM), mean structural similarity index (MSSIM), correlation coefficient (CC), and root mean square error (RMSE) \cite{chudasama2024comparison}.

\begin{figure}[!t]
  \centering
\includegraphics[width=0.42\textwidth]{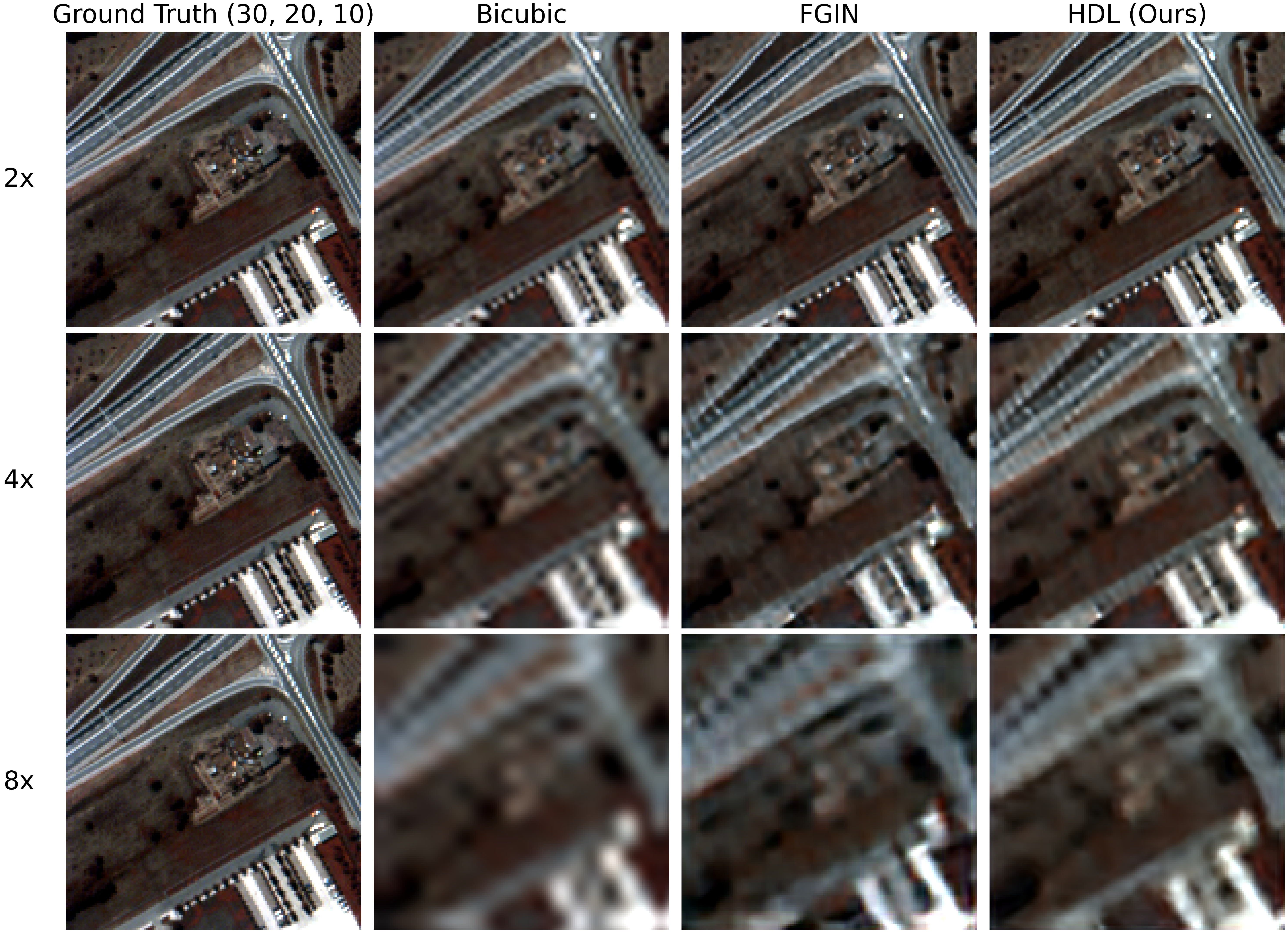}
\caption{Qualitative comparison on the PaviaU test image, representing a false-color composite, at scaling factors $2\times$, $4\times$, and $8\times$.}

  \label{fig:scales}
\end{figure}
\begin{table*}
\centering
\label{comparison}
\caption{Evaluation on datasets (PaviaC, PaviaU) in different scaling setups. The comparison results are reported from \cite{zhang2024hyperspectral}.}
\resizebox{0.67\textwidth}{!}{%
\renewcommand{\arraystretch}{0.95} 
\small 
\begin{tabular}{|c|c|ccc|ccc|}
\hline
\multirow{2}{*}{\textbf{Scale Factor}} & \multirow{2}{*}{\textbf{Model}} & \multicolumn{3}{c|}{\textbf{PaviaC}} & \multicolumn{3}{c|}{\textbf{PaviaU}} \\ 
\cline{3-8}
                      &                   & \textbf{MPSNR$\uparrow$} & \textbf{MSSIM$\uparrow$} & \textbf{SAM$\downarrow$} & \textbf{MPSNR$\uparrow$} & \textbf{MSSIM$\uparrow$} & \textbf{SAM$\downarrow$} \\ \hline
\multirow{8}{*}{\centering $\boldsymbol{2\times}$} 
    & VDSR   \cite{kim2016accurate}     & 34.87 & 0.9501 & 3.689  & 34.03  & 0.9524 & 3.258 \\ 
    & EDSR   \cite{lim2017enhanced}    & 34.58 & 0.9452 & 3.898  & 33.98  & 0.9511 & 3.334 \\ 
    & MCNet   \cite{lii2020mixed}    & 34.62 & 0.9455 & 3.865  & 33.74  & 0.9502 & 3.359 \\ 
    & MSDformer    \cite{chen2023msdformer}    & 35.02 & 0.9493 & 3.691  & 34.15  & 0.9553 & 3.211 \\ 
    & MSFMNet  \cite{zhang2021multi}           & 35.20 & 0.9506 & 3.656  & 34.98  & 0.9582 & 3.160 \\ 
    & AS3 ITransUNet    \cite{xu20233}   & 35.22 & 0.9511 & 3.612  & 35.16  & 0.9591 & 3.149\\ 
    & PDENet    \cite{hou2022deep}      & 35.24 & 0.9519 & 3.595  & 35.27  & \underline{0.9594} & \underline{3.142} \\ 
    & CSSFENet  \cite{zhang2024hyperspectral} & \underline{35.52} & \underline{0.9544} & \underline{3.542}  & \underline{35.92}  & \textbf{0.9625}  & \textbf{3.038} \\ 
    & HDL (Ours)    & \textbf{36.84} &  \textbf{0.9595} & \textbf{3.531} & \textbf{36.21} &  0.9477 & 3.538 \\ \hline
\multirow{8}{*}{\centering $\boldsymbol{4\times}$} 
    & VDSR \cite{kim2016accurate}      & 28.31     &0.7707  &6.514  & 29.90   &0.7753  &4.997 \\ 
    & EDSR  \cite{lim2017enhanced}    & 28.59  &0.7782  &6.573  & 29.89   & 0.7791 &5.074 \\ 
    & MCNet   \cite{lii2020mixed}   & 28.75   &0.7826  &6.385  & 29.99  &0.7835  &4.917 \\ 
    & MSDformer   \cite{chen2023msdformer}      & 28.81  &0.7833  &5.897  & 30.09  & 0.7905  &4.885 \\ 
    & MSFMNet  \cite{zhang2021multi}        &   28.87  & 0.7863  &6.300  & 30.28  &0.7948 &4.861 \\ 
    & AS3 ITransUNet  \cite{xu20233}  &  28.87  &0.7893  &5.972  & 30.28  & 0.7940  &4.859\\ 
    & PDENet  \cite{hou2022deep}       & 28.95  &0.7900  &5.876  & 30.29  & 0.7944 & 4.853 \\ 
    & CSSFENet   \cite{zhang2024hyperspectral}   & \underline{29.05} & \underline{0.7961} & \underline{5.816}  & \underline{30.68} & \textbf{0.8107} & \underline{4.839} \\ 
    & HDL (Ours)   & \textbf{30.08} & \textbf{0.8263} & \textbf{4.607} & \textbf{30.73} & \underline{0.8049} & \textbf{4.549} \\ \hline
\multirow{8}{*}{\centering $\boldsymbol{8\times}$} 
    & VDSR  \cite{kim2016accurate}     & 24.80   &0.4944  &7.588  & 27.02  &0.5962 &7.133 \\ 
    & EDSR   \cite{lim2017enhanced}   & 25.06   &0.5282  &7.507 & 27.46  &0.6302 &6.678 \\ 
    & MCNet  \cite{lii2020mixed}    & 25.09   &0.5391&7.429  & 27.48   &0.6254   &6.683 \\ 
    & MSDformer  \cite{chen2023msdformer}       & 25.21   &0.5462   &7.427  & 27.32  &0.6341 &6.668 \\ 
    & MSFMNet  \cite{zhang2021multi}        &   25.25   &0.5464  &7.449  & 27.58  &0.6356 &6.615 \\ 
    & AS3 ITransUNet  \cite{xu20233}  &  25.25    &0.5435&7.417  & 27.68  &0.6413 &6.574\\ 
    & PDENet   \cite{hou2022deep}      & 25.28   &0.5436  &7.402  & 27.73  & \underline{0.6457} & 6.531 \\ 
    & CSSFENet   \cite{zhang2024hyperspectral}   & \underline{25.35} & \underline{0.5493} & \underline{7.306}   & \underline{27.82} & \textbf{0.6569} & \underline{6.505} \\ 
    & HDL (Ours)        & \textbf{25.98} & \textbf{0.5964} & \textbf{5.674} & \textbf{28.16} & 0.6343 & \textbf{5.672} \\ \hline
\end{tabular}%
}
\label{tab:results2}
\end{table*}

\begin{table}[tb!]
\centering
\caption{Evaluation on the Chikusei dataset in different scaling setups. The comparison results are sourced from \cite{wang2024enhancing}.}
\resizebox{0.47\textwidth}{!}{
\begin{tabular}{|c|c|c|c|c|c|c|}
\hline
\multirow{1}{*}{\centering \textbf{Scale}} & \multirow{1}{*}{\centering \textbf{Model}} & \textbf{MPSNR$\uparrow$} & \textbf{MSSIM$\uparrow$} & \textbf{CC$\uparrow$} & \textbf{RMSE$\downarrow$} & \textbf{SAM$\downarrow$} \\ \hline
\multirow{8}{*}{\centering $\boldsymbol{2\times}$} 
    & Bicubic     & 35.008 & 0.932 & 0.965 & 0.0229 & 1.718 \\ 
    & EDSR    \cite{lim2017enhanced}    & 35.489 & 0.941 & 0.961 & 0.0198 & 2.444 \\ 
    & GDRRN   \cite{li2018single}    & 35.958 & 0.939 & 0.971 & 0.0206 & 1.561 \\ 
    & SSPSR   \cite{jiang2020learning}    & 35.723 & 0.944 & 0.965 & 0.0197 & 2.275 \\ 
    & MCNet    \cite{lii2020mixed}   & 36.371 & 0.948 & 0.971 & 0.0198 & 1.784 \\ 
    & GELIN    \cite{wang2022group}   & 37.747 & \underline{0.959} & 0.979 & 0.0170 & \textbf{1.384} \\ 
    & DIFF     \cite{wang2024enhancing}   & \textbf{38.748} & \textbf{0.966} & \underline{0.982} & \underline{0.0161} & 1.638 \\ 
    & HDL (Ours)        & \underline{38.578} & 0.956 & \textbf{0.998} & \textbf{0.0134} & \underline{1.433} \\ \hline
\multirow{8}{*}{\centering $\boldsymbol{4\times}$} 
    & Bicubic     & 29.676 & 0.770 & 0.882 & 0.0425 & 3.161 \\ 
    & EDSR    \cite{lim2017enhanced}    & 29.976 & 0.799 & 0.893 & 0.0386 & 4.127 \\ 
    & GDRRN   \cite{li2018single}     & 30.658 & 0.801 & 0.905 & 0.0374 & 2.913 \\ 
    & SSPSR    \cite{jiang2020learning}   & 30.858 & 0.823 & 0.914 & 0.0355 & 3.196 \\ 
    & MCNet     \cite{lii2020mixed}   & 31.189 & 0.821 & 0.916 & 0.0354 & 2.955 \\ 
    & GELIN     \cite{wang2022group}   & 31.095 & 0.838 & 0.914 & 0.0366 & \underline{2.834} \\ 
    & DIFF       \cite{wang2024enhancing} & \underline{32.248} & \textbf{0.860} & \underline{0.929} & \underline{0.0332} & 5.378 \\ 
    & HDL (Ours)    & \textbf{32.584} &  \underline{0.852} & \textbf{0.932} & \textbf{0.0269} & \textbf{2.220} \\ \hline
\end{tabular}
}
\label{tab:results_chikusei}
\end{table}

\subsection{Ablation Study}
To evaluate the individual contributions of each component in our proposed model, we conducted an ablation study on the PaviaU dataset with a $4\times$ upscaling factor. The results are summarized in Table~\ref{tab:ablation}, using two common hyperspectral image quality metrics such as MPSNR and SAM. The model is first trained using three residual-like blocks to establish a baseline result, following the three-layer strategy proposed in~\cite{dong2015image}. We then incorporate band grouping with a group size of $32$, as adopted by FGIN~\cite{muhammad2025fusion}. We observe that band grouping significantly improves performance on the PaviaU dataset, with MPSNR increasing from $29.84$~dB to $30.51$~dB and SAM decreasing from $5.31$ to $4.61$.

To enhance this baseline, we first integrate spectral unmixing at the early stage of the network. This addition improves both metrics, increasing MPSNR to $30.60$ and further reducing SAM to $4.49$, which highlights the importance of incorporating spectral mixing early in the network. Next, we replace the spectral unmixing with a spectral-spatial feature extraction mechanism to analyze its individual effect. This setup also improves residual learning performance, achieving $30.64$~dB MPSNR and $4.58$ SAM.

We then combine both components into a unified Spectral--Spatial Unmixing Fusion (SSUF) module followed by standard convolution. This configuration yields the best performance among MSE loss-based training, achieving $30.68$~ dB MPSNR and $4.61$ SAM. These results demonstrate the complementary benefits of combining spectral and spatial information. Further improvements are obtained by incorporating our custom loss function, which integrates spectral and spatial regularization with the MSE loss. With this loss, the model achieves the highest overall performance, reaching $30.73$~dB MPSNR and $4.54$ SAM.

We argue that while MSE minimizes per-pixel intensity differences, it does not effectively capture spectral shape variations. In contrast, the spatial--spectral gradient loss accounts for angular distortions between spectral vectors. By integrating MSE with the spatial-spectral gradient loss, the proposed loss function simultaneously addresses intensity and spectral distortions, leading to reconstructions that are both sharper and more spectrally consistent. We also investigated the impact of the model depth. Increasing the number of residual blocks from three to four in the SSUF configuration, while keeping the custom loss, leads to degraded performance ($30.51$~dB MPSNR and $4.70$ SAM), suggesting potential overfitting.

Finally, to demonstrate that SSUF can be easily adapted to any convolutional model, we replace the residual blocks with Inception-like and MobileNet blocks, as used in FGIN~\cite{muhammad2025fusion} and DSDCN~\cite{muhammad2025towards}. Integrating SSUF into these architectures significantly improves the original results, confirming its effectiveness as a plug-and-play module that generalizes well across different architectures. In addition to performance improvements, we also demonstrate significant efficiency in terms of model complexity. For instance, when SSUF is combined with three residual-like blocks, the proposed Hybrid Deep Learning (HDL) model requires only $0.33$M parameters, which is nearly three times fewer than DSDCN~\cite{muhammad2025towards}. 

A qualitative comparison on the PaviaU test image among bicubic interpolation, FGIN~\cite{muhammad2025fusion}, and our proposed HDL is also provided in Fig. 2. At the most challenging $8\times$ scale, bicubic collapses into blur; FGIN retains sharper details but exhibits slight aliasing artifacts; and HDL recovers finer edges and textures with fewer artifacts. This clearly shows that our design achieves a better balance between accuracy and efficiency, making it suitable for practical deployment in resource-constrained environments.

\subsection{Comparison with State-of-the-Art Methods}
Firstly, we evaluated the proposed Hybrid Deep Learning (HDL) model on the PaviaC and PaviaU datasets under $2\times$, $4\times$, and $8\times$ scaling setups, as shown in Table~\ref{tab:results2}. Specifically, HDL is compared with widely used models such as VDSR~\cite{kim2016accurate}, EDSR~\cite{lim2017enhanced}, MCNet~\cite{lii2020mixed}, MSDformer~\cite{chen2023msdformer}, and CSSFENet~\cite{zhang2024hyperspectral}. Across all scales, HDL consistently outperforms existing approaches, particularly in terms of MPSNR and SAM. At the $2\times$ scale, for instance, HDL achieves the highest MPSNR scores of $36.84$~dB on PaviaC and $36.21$~dB on PaviaU, along with the lowest SAM values of $3.531$ and $3.538$, respectively. Similar improvements are observed at $4\times$ and $8\times$, demonstrating the model’s strong capacity for both spatial detail restoration and spectral fidelity.

Secondly, we conducted experiments on the Chikusei dataset under $2\times$ and $4\times$ upscaling, using benchmark results from~\cite{wang2024enhancing} for comparison (Table~\ref{tab:results_chikusei}). HDL again demonstrates superior performance, especially at the more challenging $4\times$ scale, achieving the best results in MPSNR ($32.584$), CC ($0.932$), RMSE ($0.0269$), and SAM ($2.220$). These results surpass those of recent competitive methods such as DIFF~\cite{wang2024enhancing}, GELIN~\cite{wang2022group}, and MCNet~\cite{lii2020mixed}, confirming HDL’s effectiveness in handling complex hyperspectral image super-resolution tasks across different resolutions.

\section{Conclusion}
In this work, we presented HDL, a hybrid hyperspectral image super-resolution framework that integrates the Spectral-Spatial Unmixing Fusion (SSUF) module into standard 2D convolutional architectures. By combining spectral unmixing with spectral-spatial feature extraction, SSUF enhances both spatial resolution and spectral fidelity. Additionally, the use of a tailored Spatial-Spectral Gradient Loss, which jointly optimizes spatial and spectral reconstruction, enables the model to achieve robust and reliable performance across various hyperspectral scales. Experimental results on three benchmark datasets confirm that the proposed approach delivers competitive results while maintaining low model complexity, making it a practical and efficient solution for real-world remote sensing applications.

\section*{Acknowledgment}
This project has been funded by the European Union’s NextGenerationEU instrument and the Research Council of Finland under grant \textnumero{} 348153, as part of the project \emph{Artificial Intelligence for Twinning the Diversity, Productivity and Spectral Signature of Forests} (ARTISDIG). We also acknowledge CSC for awarding access to the LUMI supercomputer, owned by the EuroHPC Joint Undertaking.

\vspace{12pt}

\bibliographystyle{ieeetr}      
\bibliography{main.bbl}       

\end{document}